\documentclass{sig-alternate-05-2015}
\usepackage{booktabs}

\begin{document}

% Copyright
%\setcopyright{acmcopyright}
%\setcopyright{acmlicensed}
%\setcopyright{rightsretained}
%\setcopyright{usgov}
%\setcopyright{usgovmixed}
%\setcopyright{cagov}
%\setcopyright{cagovmixed}

% DOI
%\doi{10.475/123_4}
%
% ISBN
%\isbn{123-4567-24-567/08/06}

%Conference
%\conferenceinfo{PLDI '13}{June 16--19, 2013, Seattle, WA, USA}
%
%\acmPrice{\$15.00}

%
% --- Author Metadata here ---
% \conferenceinfo{WOODSTOCK}{'97 El Paso, Texas USA}
%\CopyrightYear{2007} % Allows default copyright year (20XX) to be over-ridden - IF NEED BE.
%\crdata{0-12345-67-8/90/01}  % Allows default copyright data (0-89791-88-6/97/05) to be over-ridden - IF NEED BE.
% --- End of Author Metadata ---

% \title{Alternate {\ttlit ACM} SIG Proceedings Paper in LaTeX
% Format\titlenote{(Produces the permission block, and
% copyright information). For use with
% SIG-ALTERNATE.CLS. Supported by ACM.}}
\title{Analyzing Wikipedia Membership Dataset and Predicting Unconnected Nodes in the Signed Networks}
% \subtitle{[Extended Abstract]
% \titlenote{A full version of this paper is available as
% \textit{Author's Guide to Preparing ACM SIG Proceedings Using
% \LaTeX$2_\epsilon$\ and BibTeX} at
% \texttt{www.acm.org/eaddress.htm}}}

% \numberofauthors{8} %  in this sample file, there are a *total*
\numberofauthors{3}
\author{
% 1st. author
\alignauthor
Zhihao Wu
% \titlenote{Dr.~Trovato insisted his name be first.}\\
      \affaddr{lawuzhihao@g.ucla.edu}\\
    %   \affaddr{1932 Wallamaloo Lane}\\
    %   \affaddr{Wallamaloo, New Zealand}\\
    %   \email{lawuzhihao@g.ucla.edu}
% 2nd. author
\alignauthor
Taoran Li 
% \titlenote{The secretary disavows
% any knowledge of this author's actions.}\\
      \affaddr{taoqi612@gmail.com}\\
%       \affaddr{P.O. Box 1212}\\
%       \affaddr{Dublin, Ohio 43017-6221}\\
    %   \email{webmaster@marysville-ohio.com}
% 3rd. author
\alignauthor 
Ray Roman 
% {\o}rv{\"a}ld\titlenote{This author is the
% one who did all the really hard work.}\\
      \affaddr{rroman681@g.ucla.edu}\\
%       \affaddr{1 Th{\o}rv{\"a}ld Circle}\\
%       \affaddr{Hekla, Iceland}\\
    %   \email{rroman681@g.ucla.edu}
\and  University of California, Los Angeles % use '\and' if you need 'another row' of author names
}
% There's nothing stopping you putting the seventh, eighth, etc.
% author on the opening page (as the 'third row') but we ask,
% for aesthetic reasons that you place these 'additional authors'
% in the \additional authors block, viz.
% \additionalauthors{Additional authors: John Smith (The Th{\o}rv{\"a}ld Group,
% email: {\texttt{jsmith@affiliation.org}}) and Julius P.~Kumquat
% (The Kumquat Consortium, email: {\texttt{jpkumquat@consortium.net}}).}
% \date{30 July 1999}
% Just remember to make sure that the TOTAL number of authors
% is the number that will appear on the first page PLUS the
% number that will appear in the \additionalauthors section.

\maketitle
\begin{abstract}
In the age of digital interaction, person-to-person relationships existing on social media may be different from the very same interactions that exist offline.
Examining potential or spurious relationships between members in a social network is a fertile area of research for computer scientists---here we examine how relationships can be predicted between two unconnected people in a social network by using area under Precison-Recall curve and ROC.
Modeling the social network as a signed graph, we compare Triadic model,Latent Information model and Sentiment model and use them to predict peer to peer interactions, first using a plain signed network, and second using a signed network with comments as context.
We see that our models are much better than random model and could complement each other in different cases.
\end{abstract}




%
%  Use this command to print the description
%

% We no longer use \terms command


\section{Introduction}
As engineering teams include more and more interactive features on social media, it sculpts the detailed interpersonal interactions between two people in cyberspace.
Several social networks such as Facebook, Wikipedia, or Stack Overflow allow users to tacitly review, critique, or applaud their peers with the use of ``Likes", ``Upvotes", ``Recommendations", or some other buzzword.
Interacting with others can be interpreted as binary, such as a ``like" or ``dislike"; they can be on some sort of scale, like an Amazon product rating from one to ten stars; or it can include extensive comments.
The nature of these links can be considered positive (``Great contribution!") or negative (``This comment appears to be irrelevant.").
Evaluations such as these form the cornerstone of signed networks, where user--user interactions can be quantified in one of these ways.
Deviating from the weighted directed graph paradigm, relationships are now represented in a multidimensional fashion \cite{wang2014case}.

Predicting the interactions between any two users in a social network need not remain a task rooted firmly in graph theory; the rich information contained in these interactions---namely the comments---provide interesting exercises in natural language processing \textsc{nlp} and sentiment analysis.
Various papers address the task of deducing a signed network based on the comments or reviews imparted from person to person \cite{DBLP:journals/corr/ElangovanE14}.
However, how well do these evaluations perform when looking at less complicated models?
We tested the rigor and performance of natural language processing in predicting pairwise user interactions.
Fitting the \textsc{nlp} methodology to the triadic and latent information models, we performed the comparison using the Wikipedia RfA data set, which is a signed network decorated with user--user evaluations.
We have discovered that the natural language processing model performs the best, while the other two models performed decently well.
% The \textit{proceedings} are the records of a conference.
% ACM seeks to give these conference by-products a uniform,
% high-quality appearance.  To do this, ACM has some rigid
% requirements for the format of the proceedings documents: there
% is a specified format (balanced  double columns), a specified
% set of fonts (Arial or Helvetica and Times Roman) in
% certain specified sizes (for instance, 9 point for body copy),
% a specified live area (18 $\times$ 23.5 cm [7" $\times$ 9.25"]) centered on
% the page, specified size of margins (1.9 cm [0.75"]) top, (2.54 cm [1"]) bottom
% and (1.9 cm [.75"]) left and right; specified column width
% (8.45 cm [3.33"]) and gutter size (.83 cm [.33"]).

% The good news is, with only a handful of manual
% settings\footnote{Two of these, the {\texttt{\char'134 numberofauthors}}
% and {\texttt{\char'134 alignauthor}} commands, you have
% already used; another, {\texttt{\char'134 balancecolumns}}, will
% be used in your very last run of \LaTeX\ to ensure
% balanced column heights on the last page.}, the \LaTeX\ document
% class file handles all of this for you.

% The remainder of this document is concerned with showing, in
% the context of an ``actual'' document, the \LaTeX\ commands
% specifically available for denoting the structure of a
% proceedings paper, rather than with giving rigorous descriptions
% or explanations of such commands.

\section{Problem Statement}
Performing detailed user--user predictions using comments and natural language processing in parallel with the topology of signed networks has already been accomplished with great results \cite{DBLP:journals/corr/0001PLP14}.
But can comparable results be achieved with simpler models, especially for an information rich network such as those centered around Wikipedia?
The forthcoming goal, then, is to predict the peer-to-peer interactions in an information network using three probabilistic soft logic models, one of which is a natural language processing (\textsc{nlp}) framework, while the other two (known as \textit{triadic} and \textit{latent information}) are rooted more firmly in sociological theory.
These three models are then compared to see if contextual information (i.e., comments) enriches the predictions, or if adequate approximations can be made without them.

To embark on this journey, the Wikipedia RfA (requests for adminship) data set is used.
This network is vast---approximately ten thousand nodes and 160 thousand edges.
Not only does this network have signed edges, but it also has comment embeddings between one user and another, summarizing person $A$'s opinion of $B$'s request for adminship.
The same methodology with the same dataset derivatives are used to perform a comprehensive comparison.

\section{Related Work}
\subsection{Pairwise Interactions in Sociology}
Uncovering the who-trusts-whom backbone within a social network can uncover mountains of information normally invisible.
Predicting pairwise relationships, positive or negative, are essential features of a network, and the ability to predict them may have great impact on marketing techniques or targeted advertising.
This task double-dips in both computer science research as well as sociological theory.
As a primer, one may consider the latter.
Ultimately, these theories take care to formulate a triangular relationship between members in a social network, yet using them as ground truths may prove to be quite inflexible, especially when considering other opinion-based features in parallel.
Thankfully, there are many flavors of person-to-person relationship predicting that relaxes these rigid constraints and provides a usable approximation.
Unfortunately, performing these computations may be rather non-trivial.

\subsection{Modeling Interactions with Probabilistic Soft Logic}
One method for approximating trust between person $A$ and person $B$ yields data not in a binary fashion, but rather on a spectrum from zero to one.
Researchers have tackled these interesting network problems by using probabilistic soft logic (\textsc{psl}) aimed at modeling the trust within a social network \cite{huang:starai12}.
Probabilistic soft logic scaffolds hard-truths with an assigned weight such as follows:
\begin{equation*}
    \textsc{trusts}(A, B) \wedge \textsc{trusts}(B, C) \overset{0.75}{\Rightarrow} \textsc{trusts}(A, C)\,,
\end{equation*}
where these key weights soften the hard logic prevalent in sociological theory.
With probabilistic soft logic, one can determine how ``truthy" something may be by its \emph{distance to satisfaction}, or plainly put, how close the program's prediction is from the actual truth \cite{huang:starai12}.
If one were to consider a set of ground truths $g \in G$, each truth $t_g$ with a weight $w_g$, then the probability distribution over this prediction is
\begin{equation*}
    Pr(x; w) = \exp \left( -\sum_{g\in G} w_g (1 - t_g(x))\right)\,.
\end{equation*}

For this kind of problem, reaching the global optimum is a challenge that may not be solved in reasonable time, especially for extremely large networks.
Therefore, a cogent solution must satisfy some sort of constraint.
Some approaches have implemented a logistic regression methodology to measure the ``goodness" of machine learning predictions compared to using sociological heuristics, while others have recast the objective function in a form more pointed towards satisfying those same sociological constraints \cite{leskovec2010predicting,DBLP:journals/corr/0001PLP14}.

\subsection{Triadic Model: Analyzing and Predicting Edges}
One obvious way to predict the edges between two peers is by a ``fill in the blank" sort of approach, by yielding a prediction that purely satisfies sociological theories.
Two of which are popular choices for the triadic model.
Familiar maxims like ``the friend of my friend is also my friend" and ``the enemy of my enemy is my friend" take the main stage of \textit{balance theory}.
For better or for worse, these truisms are framed in the context of trustworthiness between peers.
In other words, completing a \textit{triadic closure} with this theory mandates that if Amy trusts Brad, and Brad trusts Candice, then Amy must also trust Candice.

\textit{Status theory} ensures positive links are directed from subordinate to superior, and \textit{vice versa} \cite{leskovec2010signed,cosmides1992cognitive}.
This theory maintains that if Amy is in charge of Brad (a negative link), and Brad is in charge of Candice, then Amy has a negative link towards Candice.
However, the status theory cannot make a prediction when considering two members of equal social status.
If Brad is in charge of Amy and Candice, then what is the relationship between the two subordinates?
Clearly this approach has limitations when attempting to satisfy these aphorisms, and this is precisely where probabilistic soft logic can shine.
Previous work tries to evaluate the accuracy of these two social theories in static and evolving social networks \cite{leskovec2010signed}.
In brief, these theories can correctly characterize, for example, mutual friendship triads, but may underrepresent triads in balance theory.
For dynamic graphs, however, balance theory may not hold as strongly, and consequently status theory may be the key to scoring accurate predictions.

\subsection{Uncovering Discreet Assumptions in Social Networks}
Oftentimes, there is a nuance in the very fingerprint of social networks that may not normally be available at a cursory glance.
These subtleties may be considered \emph{latent information}: they are subtle, but unknown, parameters of the system that provide deeper, more structured clues about the network at hand.
Discovery, then, rests on the nodes themselves rather than the edges.
For instance, one user may have a latent characterization as ``strongly trusting of others", while another user may be considered ``moderately trustworthy"---all while these tokens of information must be learned.
However, learning these latent variables, just like learning the trust between two people, is a nontrivial task as well, since this hidden information must be learned and updated several times over the entire network.

Researchers have tried to streamline the process of learning latent variables by using what are known as hinge-loss Markov random fields, thus allowing the network to be more scalable.
Hinge-loss Markov random fields are probabilistic models where the set of features is denoted by hinge-loss functions \cite{huang2015paired}
Using a hinge-loss Markov random field framework, however, and these attributes become quite apparent, and significant performance boosts can be obtained when using the paired-dual learning algorithm for learning these hidden attributes \cite{huang2015paired}.

\subsection{Exploiting Natural Language Processing}
One may draw comparisons between the relaxed Boolean logic previously mentioned and another flavor of link prediction: sentiment analysis.
In natural language processing (\textsc{nlp}), strides have been made to project emotions and persuasions from textspace to a more program-friendly format.
The realm of sentiment analysis is a great viewpoint for modeling interpersonal relationships; researchers have even contextualized \textsc{nlp} with the same sociological theories introduced above \cite{DBLP:journals/corr/0001PLP14}.
By leveraging this framework, accomplishing the goal of guessing person-to-person relationships focuses less on discovering, for example, the identity of the social network users in question, but rather on its predictive capabilities.

But the role of \textsc{nlp} in solving these problems does not start where sociological theories end; care must be taken to ensure that the previously mentioned enshrining principles are followed \cite{DBLP:journals/corr/0001PLP14}.
However, as we have seen before, these predictions may conflict with the sociological theories previously outlined; this problem was solved by predicting edge weights not as binary, but on a spectrum \cite{DBLP:journals/corr/0001PLP14}.
In other words, the inferred sign is actually $x_e \in [0, 1]$---the same flavor of probabilistic soft logic introduced above---and thus the problem is cast as a hinge-loss Markov random field, which makes it a good approximation for an \textsc{np}-hard problem.
While much related work is fixated on determining trust rather than friendship links, this model can be extended for a signed network by mapping the transformation as $x_e = 1$ indicating a positive interaction and $x_e = 0$ indicating a negative interaction \cite{huang:starai12,huang:sbp2013,DBLP:journals/corr/0001PLP14}.
Furthermore, this sign prediction problem has been tackled in myriad forms, with some users leveraging the nature of the network itself to make predictions, while others make use of the learning algorithms and hefty dataset sizes \cite{leskovec2010predicting,Guha:2004:PTD:988672.988727}.

The methodology introduced in West et al.\ is a thorough one from which this experiment can derive \cite{DBLP:journals/corr/0001PLP14}.
Modeling the social network as a signed graph $G = (V, E, x)$ with $\left|V\right|$ vertices, $\left|E\right|$ edges of sign $x \in \{0, 1\}$, the following can be seen.
First, the authors restrict their attention to undirected graphs.
They emphasize that this restriction can be lifted, and parallel analyses can be performed on social networks where user--user interactions are implicitly directed (such as following someone on Twitter).
The authors focus primarily on triangles in the graph as this topographical feature is a natural extension of the aforementioned sociological theories.
A triangle is denoted as $\{e_1, e_2, e_3\} \subseteq E$, three edges that form a cycle, with signs $x_t = (x_{e_1}, x_{e_2}, x_{e_3}) \in \{0,1\}^3$, and $T$ is the set of all triangles in the graph.

Germane to this \textsc{nlp} problem, the objective function measures the so-called \textit{triangular polarity} of a triadic closure and characterizes it as a good approximation (i.e., close enough to the sociological theories outlined previously) or otherwise.
It takes the form
\begin{equation}
x^* = {x \in \{0, 1\}^{\left|E\right|}} \sum_{e \in E} c(x_e, p_e) + \sum_{t \in T} d(x_t)\,,
\end{equation}
where the first term describes the penalty of having each edge $x_e$ deviate from its prediction $p_e$, while the second term quantifies the discordance the triangle exhibits when viewed from the perspective of status or balance theory.
More specifically, the edge cost function $c$ is as follows:
\begin{equation*}
    c(x_e, p_e) = \lambda_1 (1 - p_e) x_e + \lambda_0 p_e (1 - x_e)\,.
\end{equation*}
The two parameters $\lambda_1, \lambda_0$ are learned by the model using hinge-loss Markov random fields via maximum-likelihood estimation.

\section{Model}
In this section, we present the three models, which are PSL model - Triadic, PSL model - Latent Information and Sentiment model.
We have already introduced the PSL, a very powerful system for probabilistic modeling using first-order logic syntax.
The key point of PSL is to find the correct rules to describe relational dependencies.
For example, in the social trust network, if Mike strongly trusts Alice, and Alice strongly trusts Jack, then we can define a rule implying that Mike will likely trust Jack.

Later, we consider the latent information provided by the dataset.
If someone obtains the most upvotes from others, it implies that this person will more likely be upvoted from his or her peers.
If somebody downvotes most of his or her peers, then it could be more possible that this person is not very active or optimistic; more likely, he or she would downvote to other candidates.
We also consider the context of each pairwise link: We think this textual information could provide more sufficient evidence to illustrate the relationships between users.
Therefore, we provide third model: the Sentiment model.

\subsection{PSL model - Triadic}
Inspired by the social science balance theory, we encode the tendency for upvoting and downvoting in our dataset using the rules listed in Figure 1.
We can see sixteen different possible rules are listed.
The meaning behind these rules are very intuitive: for example, if Mike strongly upvotes Alice, and Alice strongly upvotes Jack, then it is more likely that Mike also strongly upvotes Jack.
And if Mike strongly upvotes Alice, but Alice strongly downvotes Jack, we could infer that Mike will most likely downvote Jack because he and Alice may have the same preference. 
\begin{figure}[h]
	\centering
	\includegraphics[height=2in, width=3.5in]{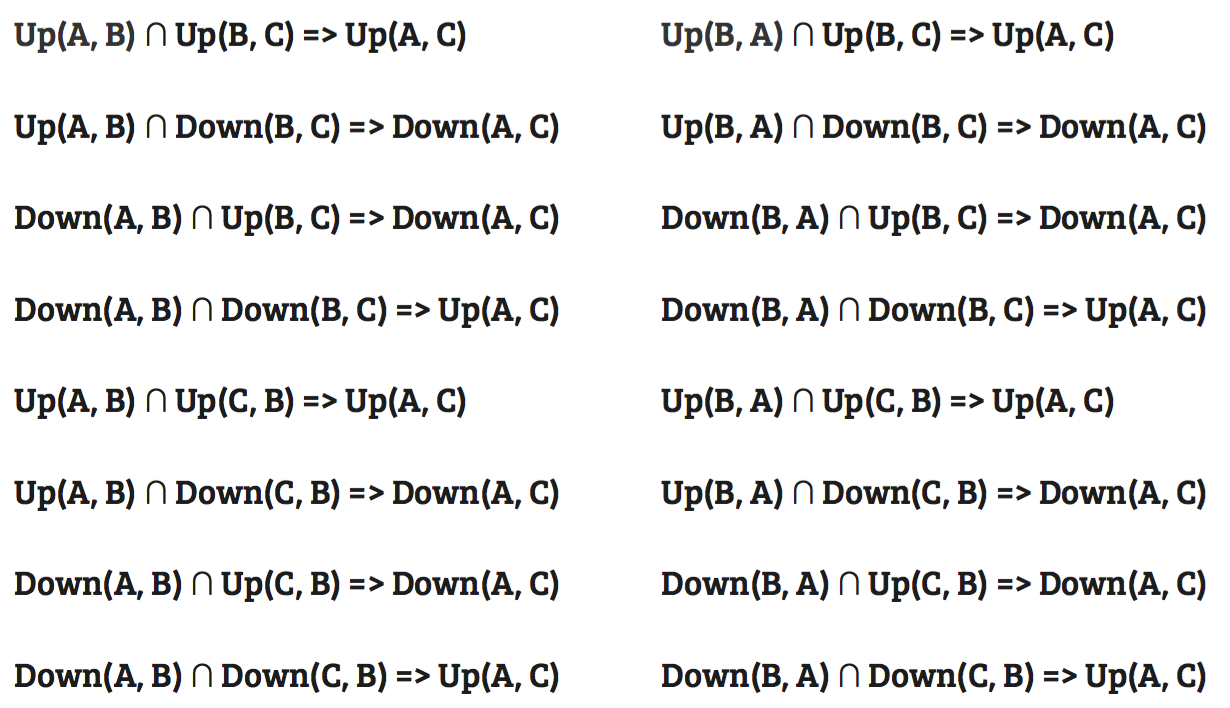}
	\caption{Rules for PSL model - Triadic. The rules are inspired from social science balance theory.}
\end{figure}

After defining these rules, we must now assign  predicted value to the unknown atoms we obtain as a set of distances from satisfaction.
An example of an atom is \textsc{Upvote}(A,B).
What is the best method to assign these values to obtain the best prediction performance?
One may approach this question from a different perspective, seeing that this is another flavor of the combinatorial optimization problem.
Particular to this model, however, is that all the link predictions are formed together, instead of one by one.
\begin{equation*}
P_r(X|w) = \frac{1}{C}\exp \left( -\sum_{r\in R} w_r (d_{r}(X))^{p}\right)\,.
\end{equation*}

In this formula, $P_r$ is the probability density distribution over assignment $X$, and $R$ is the set of rules we have defined before.
$r$ is rule's weight---in the PSL model, each rule may differentially influence the prediction. $d_r(X)$ is the rule's distance to satisfaction between $X$ and true values.
$p$ is the distance exponent, which could be linear or quadratic.
If it is quadratic, then the variation of prediction is more smooth.
Finally, $C$ is a constant. 

Notice that in our problem, the predicted value is 0 or 1, which is lends itself to binary semantics.
Here, we consider a continuous relaxation of the binary problem and find an exact nonbinary solution whose edge signs are continuous on the interval $[0, 1]$.
So as stated previously, we can cast our problem as a hinge-loss Markov random field (HL-MRF). This is inspired by \cite{bach2013hinge}, where a HL-MRF was used to predict edge signs based on balance theory.
An HL-MRF is an MRF with continuous variables and with potentials that can be expressed as sums of hinge loss terms of linear functions of variables.
HL-MRFs have the advantage that their objective function is convex so that exact inference is efficient.\cite{brocheler2012probabilistic} 

With this perspective, we can see that this problem is a maximum \textit{a posteriori} probability (MAP) estimate problem.
We train this model using various learning algorithm: inference method, maximum likelihood estimation (MLE), maximum pseudo-likelihood estimation (MPLE), and mixed method (MM) to train and test the dataset according to our model.

\subsection{PSL model - Latent Information}
The another inspiration from social science is the personality.
The previous work in signed networks focused mainly on links, whether that may be pairwise trust or friendship.
However, most of work ignored the properties that lay within the nodes.
In a trust network, for example, we can infer someone is trustworthy according to his or her profile.
More specifically, we consider additional predicates \textsc{Active} and \textsc{Favorable} in our dataset, modeling whether a member is active or favorable in (for this particular case) the Wikipedia network.
Important to note is that these predicates are \emph{not} part of the dataset, which is why this model boasts the name ``latent" information.
The intuition behind this elusive characterization is that a highly active person will more upvote other candidates, while a highly favorable candidate will receive more upvotes from his or her peers. 

While the addition of more features could potentially make the analysis more complicated, the node property could also strengthen our prediction.
On Facebook, for instance, Mike and Alice are foes, and Alice and Jack are friends.
Mike and Alice have the different political leanings: Mike leans Democrat, and Alice Republican. At the same time, Jack has Republican leanings.
In this situation, we can make prediction that Mike and Jack may bump heads more often in the political sphere than if they shared the same political leaning.
According to balance theory this prediction may be apparent, but including this extra information provides a higher degree of confidence when making the prediction.

\begin{figure}[h]
	\centering
	\includegraphics[height=2in, width=3.5in]{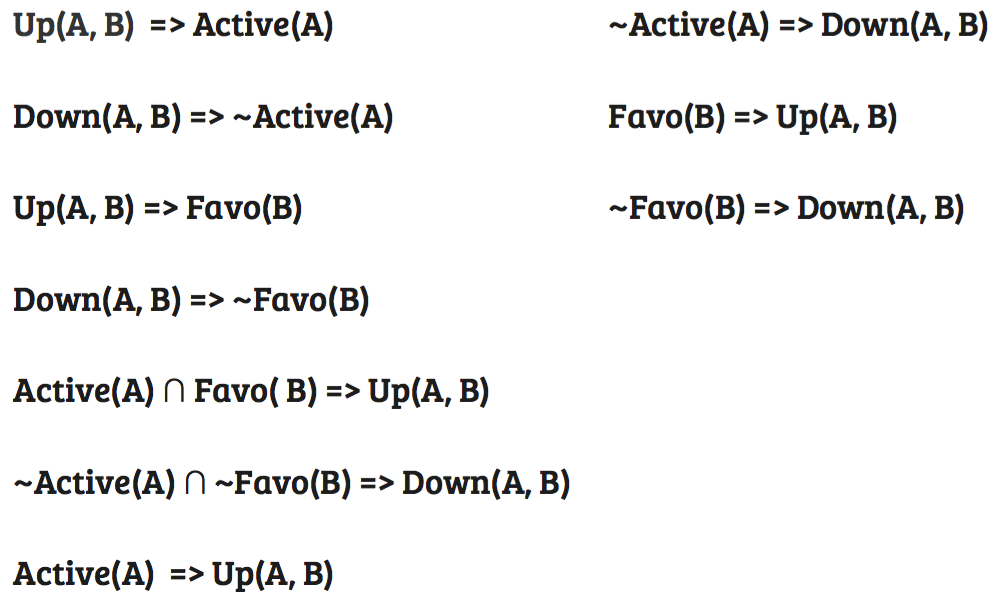}
	\caption{Rules for PSL model - Latent Information. The rules are inspired from social science.}
\end{figure}

However, we need to rewrite the PSL Triadic model to accommodate these results.
Notice that, in the previous model, we have already converted the discrete link value 0 or 1 into a continuous [0,1].
Little modification needs to be done there.
On the other hand, there are several proposed methods to solve the continuous latent variable model.
We take the probabilistic principal component analysis (PPCA) approach for this case.

Instead of using naive PCA and MLE-PCA, we use Expectation--Maximization for PCA.
The naive PCA and MLE-PCA have heavy computational taxes for high dimensional data and large (and potentially incomplete) datasets.
We rewrite the previous model formula in the PSL - triadic model. 
\begin{equation*}
\log P_r(X, Z|w) =  \sum_{n}[\log P_r(X_n|Z_n) + \log P_r(Z_n)]
\end{equation*}

Here, $Z$ represents the latent variables.
In the E step of the E--M algorithm, we compute the expectation of complete log likelihood with respect to the posterior probability of latent variables $z$ using current parameters, from which one can derive $E[z_n]$ and $E[z_{n}z_{n}^{T}]$ from posterior probability $p(z|x)$.
In the M step of the E--M algorithm, we maximize with respect to parameter $W$ and hyperparameters in the model.
This is an iterative solution, at each step updating parameters given current expectations, and expectations given current parameters. 

Although the properties that are examined include \textsc{Active} and \textsc{Favorable}, the nodes need not restrict themselves to two hidden properties.
In fact, the node of this signed network could have myriad other properties.
Selecting the best ones, then, depend on the nature of the problem being solved, so while the political position or level of education or some other parameter may be hidden, those are not as important in this problem as favorability and activity.
Obviously, considering all latent information could stifle the training process, and the computational time would be much larger with more features to train.
Our experiment in the next section also supports this intuition. 

\subsection{Sentiment model}
Real signed networks may also include many other tokens of information attached to the edges between two users, such as comments.
These comments from user $A$ to user $B$ represent the directed attitue and behavior, and thus it is highly related to the edge sign.
We continue with a sentiment model to extract the hidden sentiment information from the comments embedded in the edges and use the this treasure trove of information to make the prediction.

The sentiment model is a binary classifier which could predict whether a sign is positive or negative given the nature of the text.
Here, we use logistic regression as the binary classifier and build feature extraction to convert a comment into a input vector for the logistic regression.
First, we use bag-of-words model to extract a set of words as tokens from the text.
Then, we calculate the frequency of each token in the comments and convert the comments into a token frequency vector.
After that, we use term frequency–inverse document frequency(TF-IDF) model to evaluate how important a word is to a document.
Based on TF-IDF model, we could convert the token frequency vector into a weighted vector and use the weighted vector as the input of logistic regression.

The TF-IDF model considers term frequency and inverse document frequency together.
Term frequency $tf(t,d)$ have many a particular weighting scheme.
Here, we use the raw frequency, which is the number of times that term $t$ occurs in the document.
The inverse document frequency $idf(t,D)$ states whether the term is common or rare across all documents.
The equation is as follows: 
\begin{equation*}
    idf(t,D) = \log(\frac{N}{|{t\in D:t\in d}|})\,,
\end{equation*}
where $N$ is total number of documents in the corpus $N=|D|$.

Then, the \textsc{tf-idf} is calculated as:
\begin{equation*}
    \textsc{tf-idf}(t,d,D)=tf(t,d)\cdot idf(t,D)
\end{equation*}

Logistic regression is a common binary logistic model used to estimate the probability of a binary response based on one or more predictor variables.
Logistic regression requires a sigmoid function to predict the probability that a given item belongs to the class 1 versus the probability that it belongs to the class 0. The function is
\begin{equation*}
    P(y=1|x) = h_\theta(x)=\frac{1}{1+\exp(-\theta ^T x ) }\,,
\end{equation*}
and
\begin{equation*}
    P(y=0|x) = 1- P(y=1|x)=1-h_\theta(x)\,.
\end{equation*}
Our goal is to find appropriate $\theta$ so that the probability $P(y=1|x)$ is large for all $x$ that belongs to class 1 and small for all $x$ that belongs to class 0.
One method of approach is by using a threshold to determine to which class the given $x$ should belong.
To evaluate the prediction, logistic regression uses a cost function to evaluate the effective prediction and then it could update $\theta$ to minimize the cost function.
The cost function is as follows :
\begin{equation*}
    J(\theta) = -\sum_{i}(y^{(i)} \log(h_\theta (x^{(i)})) +(1-y^{(i)})\log(1-h_\theta (x^{(i)})))\,.
\end{equation*}
After training the logistic regression classifier with cost function, we obtain the sentiment model and then use the model to predict the sign of edges.

\section{Evalution}
\subsection{Dataset description}
The dataset we use is the Wikipedia Requests for Adminship (wiki-RfA).
This large dataset was collected by Stanford SNAP Lab and is available at \url{http://snap.stanford.edu/data/wiki-RfA.html}. 
For a Wikipedia editor to become an administrator, a request for adminship (RfA) must be submitted.
Any Wikipedia member may cast a supporting, neutral, or opposing vote.
Taking all of the users into consideration, this generates a directed, signed network in which nodes represent Wikipedia members and edges represent votes.
Particular to this dataset are comments embedded in the edges.
A positive comment example would be "a good contributor" or "I've no concern, will make an excellent addition to the admin corps".
The comments contains strong sentimental information and thus can be used as evidence to infer the edge sign.
The format of the dataset is as follows:
\begin{table}[h]
\centering
\caption{wiki-RfA Data Format}
\begin{tabular}{c l}
Key & Value\\
\hline
\texttt{SRC} & Guettarda\\
\texttt{TGT} & Lord Roem\\
\texttt{VOT} & 1\\
\texttt{RES} & 1\\
\texttt{YEA} & 2013\\
\texttt{DAT} & 19:53, 25 January 2013\\
\texttt{TXT} & clueful, and unlikely to break Wikipedia.\\
\hline\end{tabular}
\end{table}

The information we need is \texttt{SRC}, \texttt{TGT}, \texttt{VOT}, \texttt{TXT}.\texttt{SRC} is the user name of source, \texttt{TGT} is user name of target, \texttt{VOT} is the source's vote on the target, and \texttt{TXT} is the comment written by the source.
The dataset contains 10,835 nodes and 159,388 edges in total.
To build the model using this dataset, we must preprocess the dataset as needed.

First, we eliminate all ambiguous signs, leaving only positive votes and negative votes to get a purely signed network.
To get a map function from user name to a unique ID, we use a \texttt{getMap} function to map their name into a unique ID.
Then, we filter some users from the dataset, sampling 500 or 1000 nodes from the whole network to accelerate the computation.
(Given the sheer size of this dataset, sampling helps reduce the runtime drastically.)
Second, the dataset contains a sign imbalance problem.
Our dataset contains a large number of positive votes and a small number of negative votes; 
in reality, we are more interested in the negative votes since they reveal rich emotional and sentiment information that contravenes any sort of optimism normally found in votes.
To make the problem more interesting, we filter part of the positive signs and keep all the negative signs to make sure that there is a steady balance of both.
After preprocessing, the dataset could be used as the input of the data pipeline.

\subsection{Evalution metrics}
\subsubsection{Evidence ratio}
In our dataset, we only know part of the edge signs of the whole network.
To train and test our models, we must partition the data into some for training and some for testing.
All the training data is considered as observed data, which implies they don't need to be inferred.
The evidence ratio is the ratio of the size of training data to the size of test data.
We hope to get better prediction as we increase the size of the training dataset.

\subsubsection{Metrics}
In information retrieval, precision is a measure of result relevancy, while recall is a measure of how many truly relevant results are returned.
Precision ($P$) is defined as the number of true positives ($T_p$) over the number of true positives plus the number of false positives ($F_p$):
\begin{equation*}
    P=\frac{T_p}{T_p + F_p }\,.
\end{equation*}
It is a measure of the fraction of true positives over all positively-predicted items.

Recall ($R$) is defined as the number of true positives ($T_p$) over the number of true positives plus the number of false negatives ($F_n$):
\begin{equation*}
    R=\frac{T_p}{T_p + F_n }\,.
\end{equation*}

Although precision and recall could be used as metrics, by themselves they are too weak to appropriately measure the classifier.
Instead, we introduce the Receiver Operating Characteristic (ROC) curve and use area under the curve (AUC) as our final metric.
The ROC curve, is a graphical plot that illustrates the performance of a binary classifier system as its discrimination threshold is varied.
It uses the true positive rate as the $y$-axis and the false positive rate as the $x$-axis.
We denote the area under the ROC curve as $\text{AUC} / \text{ROC}$. 
Importantly, $\text{AUC} / \text{ROC}$ may ignore the prediction quality of negative edges if they are rare; to remedy this, we introduce the area under the Precision--Recall curve as another metric, we denote it as $\text{AUC}/\text{negPR}$.

\subsection{Model implementation}
\subsubsection{PSL Model implementation}
Probabilistic soft logic (PSL) is a machine learning framework for development probabilistic models.
In PSL, we could define our model using a straightfoward logical syntax and solve them with fast convex optimization.
PSL is a logic programming language, so we use the logic to express the statistical relational learning problem.
It contains three concepts: Predicate, Atom and Rule.
We build our triadic model and latent information model based on different logic rules.
The predicate is used to represent a certain relationship or property like $\textsc{Friends}(A,B)$.
An atom is used to represent random variable like $\textsc{Friends}(\text{Steve}, \text{Jay})$.
Finally, a rule is used to capture the dependency or constraint like 
\begin{equation*}
    3.0:\textsc{Friends}(A,B) \quad \& \quad  \textsc{Friends}(B,C) \rightarrow \textsc{Friends}(A,C)\,.
\end{equation*}
The value prefix represents the weight this rule holds, and this value could be tuned from a learning progress.

In the training process, we define $N$ as the number of folds.
$N$ has two meanings in our training process.
First it denotes the number of partitions we split the data.
For each time, we use $N - 1$ folds as the training set and use the remaining one fold as the test set.
$N$ may also represent the number of times cross validation is used.
We train and predict for $N$ folds and in each fold, we will choose a different fold as test set and the remaining folds as training set.
So a larger $N$ means not only our evidence ratio is larger but also our cross validation will take more time.
Because of limited time for our project, we only choose the value $N$ as 3, 6 and 9.
The reason of using $N$ folds like that is because we want to make use of our known edges, so we put each known sign either into training set or test set.

\subsubsection{Sentiment model implementation}
We used the Python \texttt{scikit-learn} library to build a data pipeline and add logistic regression classifier into the pipeline.
\texttt{scikit-learn} already provides a \texttt{feature\_extraction} library.
We use \texttt{CountVectorizer} to convert the input text into token frequency vector and then the \texttt{tf\_vectorizer} to convert the token frequency vector into a weighted vector.
Next, we split the input dataset into training set and test set using \texttt{cross\_validation}.
We trained a logistic regression classifier and used it to predict the test set.
After obtaining the probability, we called \texttt{roc\_auc\_score}, \texttt{average\_precision\_score}, two functions from \texttt{sklearn.metrics}, and then we obtained the prediction results like $\text{AUC} / \text{ROC}$ and $\text{AUC}/\text{negPR}$.

\section{Results}
In this section, we demonstrate the prediction quality of different models and compare their performance in different cases.
\begin{figure}[h]
\centering
\includegraphics[height=2in, width=6in]{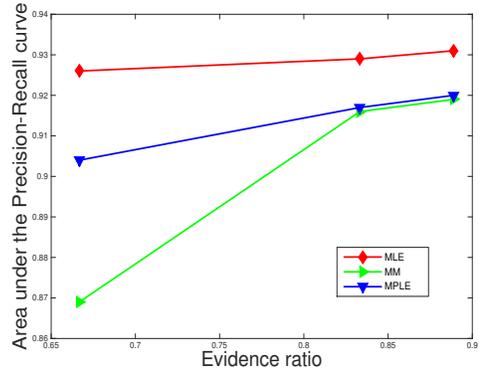}
\caption{Prediction accuracy($\text{AUC}/\text{posPR}$) from the Triadic model trained with different estimation methods on small dataset.}
\label{fig:r1}
\end{figure}

As we can see in Figure~\ref{fig:r1}, $\text{AUC}/\text{posPR}$ could achieve better score as the evidence ratio increases.
The PSL model provides three methods of estimation to evaluate the score when we train the PSL model.
We find that the Maximum Likelihood Estimation (MLE) function could tune the best parameters for our model in most cases, so we use MLE as our estimation when we train our models.

\begin{figure}[!htbp]
\centering
\includegraphics[height=2in, width=6in]{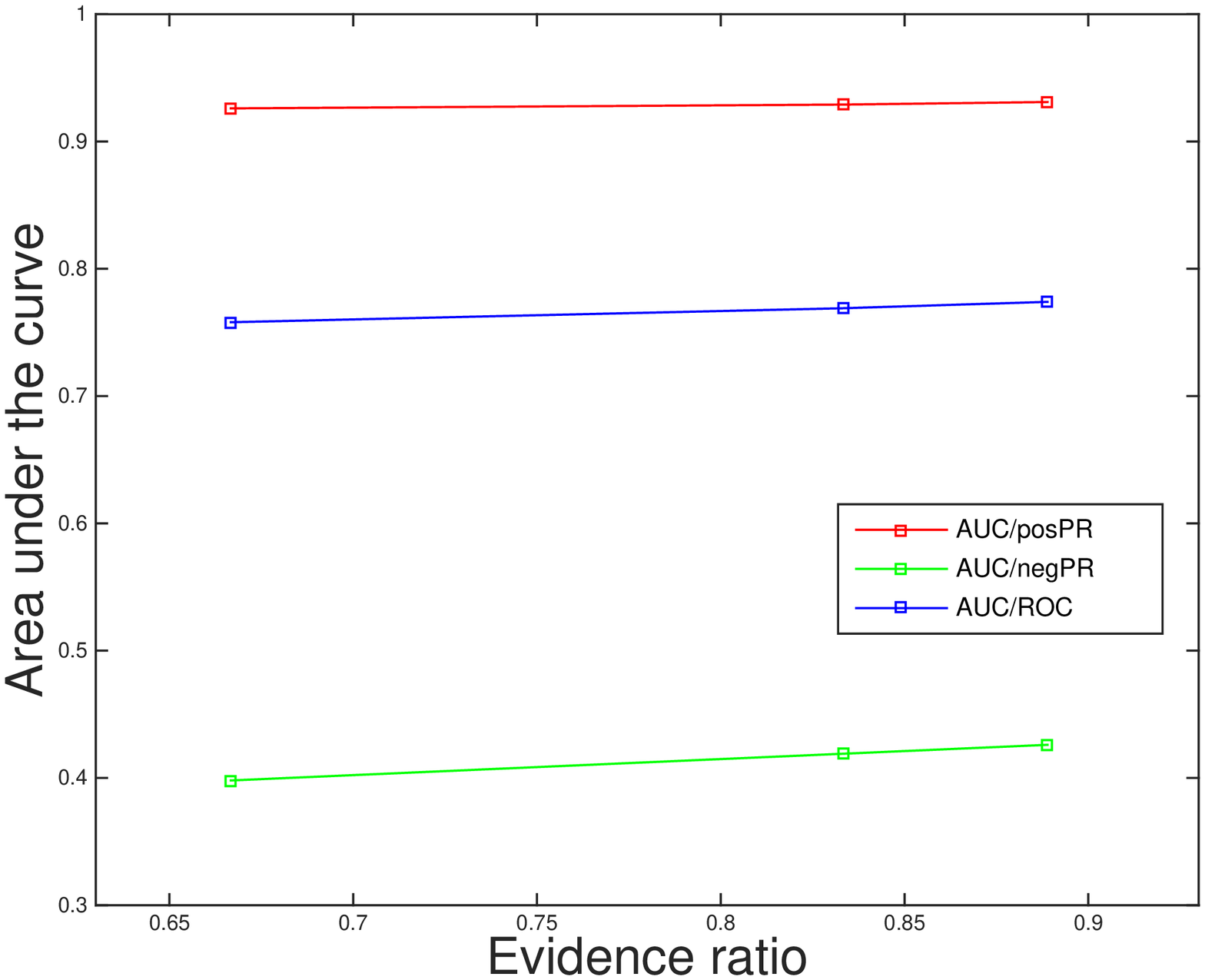}
\caption{Prediction accuracy from Triadic model under different AUC metrics  on small dataset}
\label{fig:r2}
\end{figure}

In Figure~\ref{fig:r2}, we use several metrics to measure prediction accuracy.
As you can see, $\text{AUC}/\text{negPR}$ shows the worst results.
In fact, $\text{AUC}/\text{negPR}$ is a better indicator that can show the prediction quality of negative signs if the proportion of negative sign is too small.
So we mainly focus on this metric, and our Triadic model could achieve 0.4 score: better than the score of a random guess model.

\begin{figure}[!htbp]
\centering
\includegraphics[height=2in, width=6in]{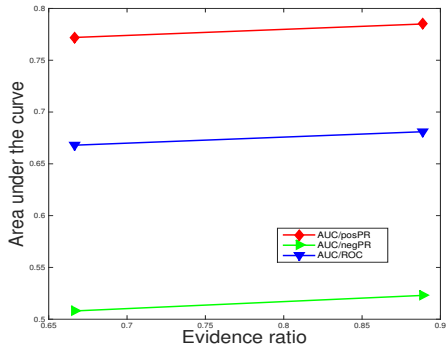}
\caption{Prediction accuracy from Triadic model under different AUC metrics on sampled dataset}
\label{fig:r3}
\end{figure}

In Figure~\ref{fig:r3}, to improve the $\text{AUC}/\text{negPR}$, we sampled the data to add more negative signs and thus minimize the sign imbalance problem.
After using sampling method, the $\text{AUC}/\text{negPR}$ score reached to 0.5: better than the previous result.

\begin{figure}[!htbp]
\centering
\includegraphics[height=2in, width=6in]{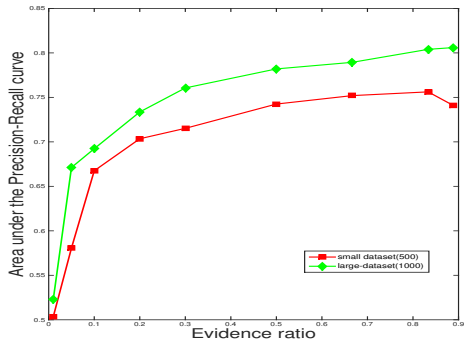}
\caption{Prediction accuracy($\text{AUC}/\text{negPR}$) from the sentiment model on small dataset and large dataset}
\label{fig:r4}
\end{figure}

In Figure~\ref{fig:r4}, we show the prediction quality of the Sentiment model.
As can be seen, the Sentiment model itself is much better as it requires relatively less data to get better results.

\begin{table}[!htbp]
\centering
\caption{Prediction accuracy from the Latent Information model under different AUC metrics  on small datase}
\label{tab:t2}
\begin{tabular}{lll}
Latent Information & MLE & MPLE\\ \hline
$\text{AUC}/\text{posPR}$ & 0.833/1.533 & 0.869/1.828\\
$\text{AUC}/\text{negPR}$ & 0.183/2.203 & 0.294/1.548\\
$\text{AUC}/\text{ROC}$ & 0.501/2.951 & 0.605/3.971\\ 
\hline\end{tabular}
\end{table}

In Table~\ref{tab:t2}, we show the results of Latent Information model.
This model did not give us any reliably good results.
It appears adding latent information to our Triadic model doesn't improve the predictions, much to our consternation.
We consider two reasons that may cause the problem:
First, the \texttt{wiki-RfA} data set does not contain strong latent in formation in users' voting, so the latent model just does not work as well as it could have.
Second, our training data set is not large enough.
Because the Latent Information model needs more relationship information for each pair of users,  each user (ideally) should keep a minimum number of neighbors to hold enough latent information that can be learned.
While this could have been ameliorated with better sampling, the primary concern was minimizing training time cost, and as such we were unable to provide a large enough training set for our model.

\section{Conclusions}
In this paper, we proposed three models to solve the sign prediction problem in signed network. First, we present the Tradic model, which inspired by the balance theory from social science. In this model, we created the sixteen different rules according to the structural theory and define two atoms like upvote and downvote. Second, we considered the personality in the signed network as latent information. Therefore we add two additional predicates Active and Favorable and ten new different rules about these two predicates. Third, because the dataset we analyzed has textual context in the form of comments from other votees, we utilized this information and built sentiment model to make predication. 

We also compared the performance of Triadic model, Latent Information model and Sentiment model, and we have shown that they are much better than a random-guess model.
We find that Sentiment model give us better results though it requires some additional contextual data.
We also find that Triadic model could make use of the network information and provides us some useful prediction.
These two model may even complement each other in different cases.
Although our Latent Information model did not work very well, we believe the model does capture hidden information and could be useful in other cases.

In future work, we may look for a better way to combine the sentiment model into PSL model, which means that we can define several new rules that relies on the implied comment information in the signed networks. The intuition is that by computing the similarity of two Wikipedia member according to their communication content, it adds similarity into PSL model and make the predictions. 
% This paragraph will end the body of this sample document.
% Remember that you might still have Acknowledgments or
% Appendices; brief samples of these
% follow.  There is still the Bibliography to deal with; and
% we will make a disclaimer about that here: with the exception
% of the reference to the \LaTeX\ book, the citations in
% this paper are to articles which have nothing to
% do with the present subject and are used as
% examples only.
%\end{document}  % This is where a 'short' article might terminate

%ACKNOWLEDGMENTS are optional
\section{Acknowledgments}
First, We thank Professor Sun for facilitating our research. Professor Sun gave us lots of valuable suggestions and helped us find an interesting and challenging topic. Also, we should thank to Stanford Network Analysis Project (SNAP), which provides many useful data set.
Finally, we also want to thank PSL programmers who are kind to answer our naive questions at Google Groups.

\bibliographystyle{abbrv}
\bibliography{sig-alternate-sample}  % sigproc.bib is the name of the Bibliography in this case
% ACM needs 'a single self-contained file'!
%\balancecolumns
% \appendix
% %Appendix A
% \section{Headings in Appendices}
% The rules about hierarchical headings discussed above for
% the body of the article are different in the appendices.
% In the \textbf{appendix} environment, the command
% \textbf{section} is used to
% indicate the start of each Appendix, with alphabetic order
% designation (i.e. the first is A, the second B, etc.) and
% a title (if you include one).  So, if you need
% hierarchical structure
% \textit{within} an Appendix, start with \textbf{subsection} as the
% highest level. Here is an outline of the body of this
% document in Appendix-appropriate form:
% \subsection{Introduction}
% \subsection{The Body of the Paper}
% \subsubsection{Type Changes and  Special Characters}
% \subsubsection{Math Equations}
% \paragraph{Inline (In-text) Equations}
% \paragraph{Display Equations}
% \subsubsection{Citations}
% \subsubsection{Tables}
% \subsubsection{Figures}
% \subsubsection{Theorem-like Constructs}
% \subsubsection*{A Caveat for the \TeX\ Expert}
% \subsection{Conclusions}
% \subsection{Acknowledgments}
% \subsection{Additional Authors}
% This section is inserted by \LaTeX; you do not insert it.
% You just add the names and information in the
% \texttt{{\char'134}additionalauthors} command at the start
% of the document.

\end{document}